\documentclass[a4paper,conference]{IEEEtran}
\ifCLASSINFOpdf
\else
\fi
\usepackage{algorithm}
\usepackage{algorithmic}
\usepackage{physics}
\usepackage{amsfonts}
\usepackage{booktabs}
\usepackage{placeins}
\usepackage{mathtools}
\usepackage{hyperref}

\makeatletter
\newcommand{\printfnsymbol}[1]{%
  \textsuperscript{\@fnsymbol{#1}}%
}

\makeatother

\newcommand\blfootnote[1]{%
  \begingroup
  \renewcommand\thefootnote{}\footnote{#1}%
  \addtocounter{footnote}{-1}%
  \endgroup
}
\begin{document}
\title{SHERLock: Self-Supervised Hierarchical Event Representation Learning}

\author{\IEEEauthorblockN{S Roychowdhury$^{*}$}
\IEEEauthorblockA{Indian Institute of Technology, Kharagpur}
\and
\IEEEauthorblockN{S A Sontakke$^{*}$\\L Itti}
\IEEEauthorblockA{University of Southern California}
\and
\IEEEauthorblockN{M Sarkar,  M Aggarwal, P Badjatiya\\ N Puri, B Krishnamurthy}
\IEEEauthorblockA{Adobe}}
% \and
% \IEEEauthorblockN{Laurent Itti}
% \IEEEauthorblockA{University of Southern California}}

%Telephone: (800) 555--1212, Fax: (888) 555--1212}


\maketitle
\def\approachName{SHERLock\space}
\begin{abstract}
\label{sec:abstract}
Temporal event representations are an essential aspect of learning among humans. They allow for succinct encoding of the experiences we have through a variety of sensory inputs. Also, they are believed to be arranged hierarchically, allowing for an efficient representation of complex long-horizon experiences. Additionally, these representations are acquired in a self-supervised manner. Analogously, here we propose a model that learns temporal representations from long-horizon visual demonstration data and associated textual descriptions, without explicit temporal supervision. Our method produces a hierarchy of representations that align more closely with ground-truth human-annotated events (+15.3) than state-of-the-art unsupervised baselines. 
Our results are comparable to heavily-supervised baselines in complex visual domains such as Chess Openings, YouCook2 and TutorialVQA datasets. Finally, we perform ablation studies illustrating the robustness of our approach. We release our code and demo visualizations in the Supplementary Material. 
\end{abstract}


% \fi
\IEEEpeerreviewmaketitle

\section{Introduction and Related Work}
\begin{figure}[t]
    \centering
    \includegraphics[width=\linewidth]{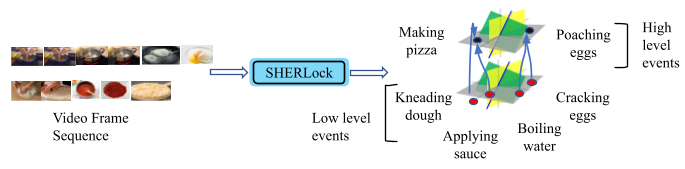}
    \caption{Overview of our approach. SHERLock learns a hierarchical latent space of events describing long horizon tasks like cooking using tutorial video and associated textual commentary.}
    \label{fig:ApproachOverview}
\end{figure}
\blfootnote{$*$ Equal Contribution}
% Offline learning has generated substantial interest recently, with efforts being applied to utilize large offline demonstration datasets to improve upon traditional \emph{tabula rasa} learning. 
Multi-modal Event Representation learning in demonstration videos is a challenging problem. Existing methods attempt to chunk videos into events using prohibitively expensive, heavily annotated datasets containing labels for objects per frame and timestamps for activities in the video. Further, existing methods suffer from sub-optimal performance when learning event representations for long sequences. In contrast, humans excel in such scenarios - given a video demonstration (Figure~\ref{fig:ApproachOverview}) of a complex task (such as cooking), humans can subconsciously abstract events (such as boiling, frying, pouring, etc.) that succinctly encode sub-sequences in such demonstrations~\cite{pammi2004chunking}. These events are hierarchical in nature~-~lower-level events are building blocks for higher-level events~\cite{naim2019emergence}. 
%We posit that this hierarchy allows humans to learn in a few-shot manner.  

% We attempt to discover such a hierarchy of concepts from demonstration data available for complex long horizon tasks such as cooking and chess gameplay. Discovering such concepts automatically from demonstration data is a non-trivial problem. 
To learn such hierarchical events, we propose an end-to-end trainable Seq2Seq architecture, SHERLock (Self-supervised Hierarchical Event Representation Learning), for multi-modal hierarchical representation learning from demonstrations. 
SHERLock takes a long-horizon sequence of demonstration images (in our case, chess, tutorial, and cooking) and commentary as input. It can then isolate semantically meaningful subsequences in input trajectories. Through ablations, we show how variants of SHERLock discover meaningful subsequences using only a sequence of images.  Our method \emph{does not require} timestamps of video and commentary, nor does it need any alignment annotation between video and textual inputs. We only assume the order of the events and narration are preserved in the input data. 
Our architecture discovers event representations along with their hierarchical organization without any supervision.  
% Our work extends the architecture in \cite{Shankar2020Discovering} to simultaneously discover event representations along with their hierarchical organization, without any supervision.
% \cite{shankar2019discovering} introduces a sequence-to-sequence architecture that clusters long-horizon action trajectories into shorter temporal skills. 
% However, their approach treats skills as independent concepts. 
% In contrast, humans organize these concepts in hierarchies where lower-level concepts can be grouped to define higher-level concepts. 

% We emphasize that it does not require temporal annotations which link subsequences in the trajectories of images to the free-flowing commentary, but instead, autonomously discovers this mapping. Utilizes multimodality - text and speech for interpretability, deals with temporality for long horizon, concepts are acquired completely offline without access to skills, hierarchical in nature improving overlap. 

% Therefore, this work takes a step towards unsupervised video understanding of high-dimensional data.    
SHERLock improves upon the state-of-the-art of related work in the following ways:
\begin{enumerate}
\item\textbf{Self-supervised:} State-of-the-art approaches in allied fields \cite{chen2019towards, boggust2019grounding, tosi2020distilled, end2end, actbert, videobert} (skill learning, event detection etc.) require large datasets of demonstrations, with expensive human annotations for timestamps corresponding to each event. \cite{Shankar2020Discovering} discover motor primitives from demonstrations but in a non-hierarchical fashion. SHERLock, on the other hand, abstracts hierarchical event representations from multimodal data, i.e. it divides 
long-horizon trajectories into a hierarchy of semantically meaningful subsequences, without requiring any temporal annotations.  

% \leavevmode\newline
\item\textbf{Long Horizon:} Long-horizon tasks remain the bane of learning systems, due to an aggregation of sub-optimal behavior over a horizon \cite{nicolescu2003natural}. Previous works in imitation learning \cite{schaal1997learning, esmaili1995behavioural, pastor2009learning, niekum2012learning, atkeson1997robot, peters2013towards}, show how agents can learn representations for events in simple tasks like cart-pole from demonstrations. More recently, \cite{schmeckpeper2019learning} shows that agents can learn action representation using a large corpus of observation data, i.e., trajectories of states and a relatively smaller corpus of interaction data, i.e., trajectories of state-action pairs. However, these approaches all restrict themselves to short horizons, while SHERLock is able to generate meaningful event representations for long-horizon tasks like cooking and chess.   

\item\textbf{Offline abstraction:} Recent works in unsupervised skill/event discovery \cite{eysenbach2018diversity,sharma2019dynamics,xu2018neural,huang2019neural} require costly interactions with an environment to discover skill sequences, an infeasible assumption in domains such as cooking or healthcare, where exploration is potentially dangerous. \cite{eysenbach2018diversity} learn a large number of low-level sequences of actions by forcing the agent to produce skills that are different from those previously acquired. Similarly, \cite{sharma2019dynamics} attempt to learn skills such that their transitions are almost deterministic in a given environment. However, these approaches require access to an environment while  SHERLock discovers these representations from offline demonstrations, utilizing large amounts of demonstration data. 
    
\item\textbf{View Invariance:} SHERLock abstracts events from demonstrations of a variety of cooking tasks. The demonstration videos originate from a number of sources, varying in camera angles, instructional styles, etc. Recent works in unsupervised skill/event learning are more restrictive \cite{Shankar2020Discovering, eysenbach2018diversity,sharma2019dynamics}, requiring that demonstration data originate from a single viewpoint, with coincident state and action spaces. 
    
\item\textbf{Multi-modality and Interpretability:} SHERLock learns a joint latent space for events utilizing both textual and visual inputs which are available in typical human demonstrations. This allows us to both visualize the physical manifestation of a temporal event, and describe in words the outcome. This is an improvement upon recent works in unsupervised skill learning, which utilize demonstrations corresponding to low dimensional state spaces and simple control signals \cite{peters2013towards, niekum2012learning, xu2018neural, huang2019neural}.  
    
\item\textbf{Hierarchical:} We find that hierarchical events abstracted by SHERLock are indeed more semantically meaningful and align more closely with ground-truth annotations for events in real-world datasets (YouCook2 \cite{ZhXuCoCVPR18} , Chess Opening and TutorialVQA \cite{colas2019tutorialvqa}) than other non-hierarchical approaches \cite{Shankar2020Discovering}. See \textit{Non-Hierarchy} in Table~\ref{tab:baseline}.
\end{enumerate}
% \end{itemize}
\section{Approach}
\label{sec:approach}
\begin{figure*}[t]
    \centering
    \includegraphics[width=0.9\linewidth]{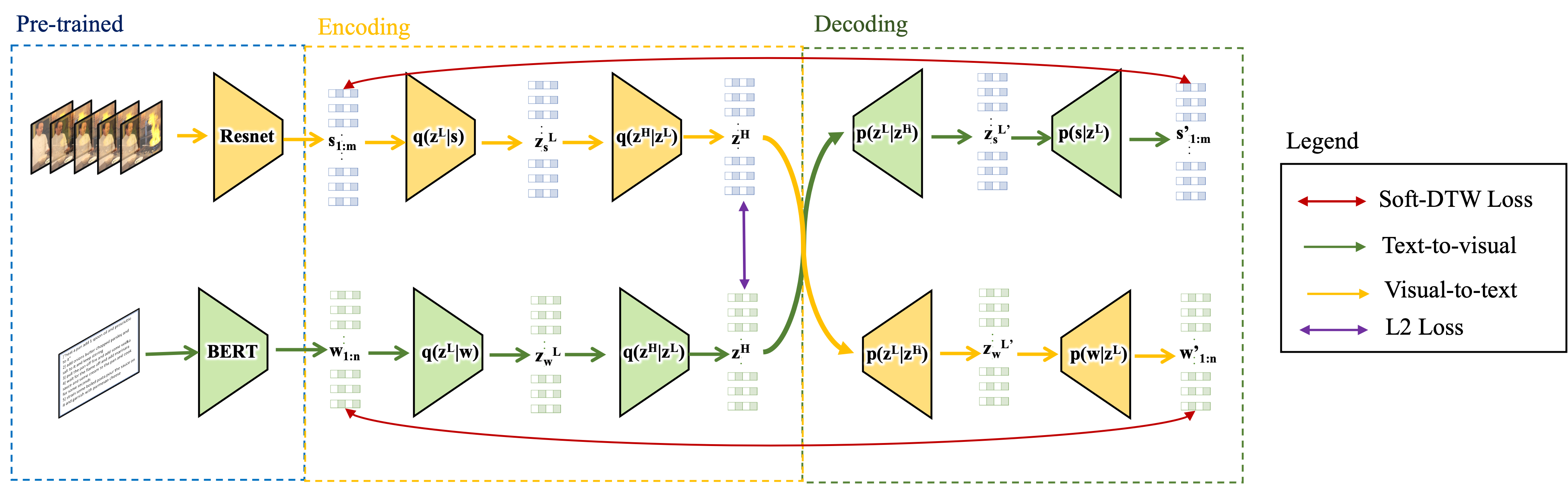}
    \caption{Overview of our approach. SHERLock learns a semantically-meaningful hierarchical embedding space which allows it to perform complex downstream tasks such as temporal event segmentation and label prediction.
    We start by encoding video and text streams into latent representations separately ($\vb{s}_{0:n}$ and $\vb{w}_{0:m}$). These are then further encoded into a hierarchy, currently consisting of low-level ($\vb{z}^L$) and high-level ($\vb{z}^H$) events. 
    During training, we swap the representation hierarchies between video and text, such that the training loss Soft-dtw($\vb{s}_{0:m},\vb{s}'_{0:m}$) and Soft-dtw($\vb{w}_{0:n},\vb{w}'_{0:n}$) will align both representations.}
    \label{fig:net_architecture}
\end{figure*}
%In this section, we provide details of our proposed approach in Section~\ref{sec:approach:unhcle} along with the details about the learning objective used for our approach in Section~\ref{sec:approach:eval_metrics}. In Section~\ref{sec:approach:eval_metrics} we provide details about the evaluation metrics used and additionally propose a new metric for improved evaluation.
\subsection{Overview}
We explain the motivation for SHERLock with an example in the domain of cooking demonstrations. Consider a long horizon demonstration for example, of an Eggs Benedict recipe. Here, low-level events might include boiling water or addition of eggs to water. Several such low-level events may combine to produce a high-level event - e.g., poaching an egg, which consists of boiling water, addition of egg to water, and finally removal after two minutes of cooking. SHERLock learns embeddings for such low and high-level events.

Broadly, SHERLock (Figure \ref{fig:net_architecture}) can be described as a multi-modal, hierarchical, sequence-to-sequence model. The model receives as input a sequence of pre-trained ResNet-50 Embeddings (\cite{he2016deep}), in addition to a sequence of pre-trained BERT-base \cite{devlin2018bert} embeddings. The two modalities are encoded separately by two transformer models into a pair of sequences of low-level latent event embeddings (e.g., boiling water or placing eggs in water, derived from either video or text). Such low-level sequences are further encoded by another pair of transformers that generate sequences of high-level event embeddings (e.g., poaching an egg). The embedding pairs are aligned through an L2 loss, forcing both representations to correspond to one another. Subsequently, a cross-modal decoding scheme is implemented: visual embeddings are used to re-generate word / BERT-base embeddings, while textual embeddings are used to generate video frame ResNet embeddings. After successful training, the system hence is expected to generate modality and domain invariant embeddings for temporal events. Those embeddings could subsequently be used for event classification and robotic skill learning (to be developed in future work).
\subsection{Hierarchical Events}
\label{sec:approach:unhcle}
Intuitively, we define an event as a short sequence of states which may occur repeatedly across several demonstration trajectories.
Events have an upper limit on their length in time steps.
They can be obtained from both a sequence of demonstration images ($\vb{S} = \vb{s}_{0:m}$) and from the associated textual description ($\vb{W} = \vb{w}_{0:n}$). 
Additionally, they are hierarchical in nature - thus, low-level and high-level events representations are denoted by $\vb{z}^L$ and $\vb{z}^H$, respectively (while the following discussion is restricted to two levels, we explore the effect of more levels in Table~\ref{tab:ablation}).
Given a low-level event representation, an associated sequence (of words or images) can be obtained using a decoder $\Phi^{x-dec}$:
\begin{equation}
\small
    \begin{split}
        \vb{x}_t | \vb{z}_t^L \sim \mathcal{N}(\vb{\mu}_{x,t}, \vb{\sigma}^2_{x,t})\\\text{where}~[\vb{\mu}_{x,t}, \vb{\sigma}^2_{x,t}] = \Phi^{x-dec}(\vb{z}_t^L,\vb{x}_{\leq t-1})
    \end{split}
\end{equation}
where $\vb{X} = \vb{x}_{0:T}$ may correspond to the flattened embedding of words $\vb{W}$ or images $\vb{S}$, and $\mathcal{N}(\cdot|\cdot)$ is a Gaussian distribution (assume prior) with parameters generated by the neural network $\Phi^{H-dec}$. 
Events also exhibit a temporal hierarchy. High-level events are generated as:
\begin{equation}
\begin{split}
    \vb{z}_t^H | \vb{z}_{\leq t-1}^H \sim \mathcal{N}(\vb{\mu}_{H,t}, \vb{\sigma}^2_{H,t})\\ \text{where}~[\vb{\mu}_{H,t}, \vb{\sigma}^2_{H,t}] = \Phi^{H-dec}(\vb{z}_{\leq t-1}^H)
\end{split}
\end{equation}
 Given such a high-level event $\vb{z}_t^H$, the associated sequence of low-level events can be approximated through a function $\Phi^{L-dec}$ as:
\begin{equation}
\begin{split}
    \vb{z}_t^L | \vb{z}_t^H,\vb{z}_{\leq t-1}^L \sim \mathcal{N}(\vb{\mu}_{L,t}, \vb{\sigma}^2_{L,t})\\ \text{where}~[\vb{\mu}_{L,t}, \vb{\sigma}^2_{L,t}] = \Phi^{L-dec}(\vb{z}_t^H,\vb{z}_{\leq t-1}^L)
\end{split}
\end{equation}
Thus, the resulting joint model mapped over trajectories of images $p(\vb{S}, \vb{z}^L, \vb{z}^H)$ factorizes as:
\begin{equation}
\begin{split}
    p(\vb{s}_0)\prod_{t=1}^{m}p(\vb{s}_t|\vb{z}^L_{\leq t},\vb{s}_{<t})p(\vb{z}_t^L|\vb{z}_{<t}^L, \vb{z}_{<t}^H)p(\vb{z}_t^H|\vb{z}_{<t}^H)
\end{split}
\end{equation}
and the resulting joint model mapped over trajectories of words $p(\vb{W},\vb{z}^L, \vb{z}^H)$ factorizes as:
\begin{equation}
\begin{split}
    p(\vb{w}_0)\prod_{t=1}^{n}p(\vb{w}_t|\vb{z}^L_{\leq t},\vb{w}_{<t})p(\vb{z}_t^L|\vb{z}_{<t}^L, \vb{z}_{<t}^H)p(\vb{z}_t^H|\vb{z}_{<t}^H)
\end{split}
\end{equation}
% The functions $\Phi^{x-dec}$, $\Phi^{L-dec}$ and $\Phi^{H-dec}$ are approximated by sequence-to-sequence models (in this case transformers \cite{vaswani2017attention}). 
The transition functions $p(\vb{z}_t^L|\vb{z}_{<t}^L, \vb{z}_{<t}^H)$ and $p(\vb{z}_t^H|\vb{z}_{<t}^H)$ are also learned using fixed length transformer models \cite{vaswani2017attention}.     

\subsection{Architecture}
SHERLock consists of 2 pairs of encoding transformers - one pair for each of the modalities. For a modality $X$, where $X\in \text{set of modalities }M=\{\text{images}, \text{words}\}$, the pair of encoders consists of $q(\vb{z}_x^L|\vb{X})$, which encodes the modality $X$ into low-level events $\vb{z}_x^L$ and $q(\vb{z}_x^H|\vb{z}_x^L)$ which encodes low-level events $\vb{z}_x^L$ into high level events $\vb{z}_x^H$. 
\begin{equation}
    \vb{z}_x^L=q(\vb{z}_x^L|\vb{X})~\text{and}~\vb{z}_x^H=q(\vb{z}_x^H|\vb{z}_x^L)
\end{equation}
Analogously, SHERLock also contains 2 pairs of decoding transformers - one pair for each of the modalities. Decoding occurs in a cross-modal manner - textual events generate video and visual events generate text. Thus, for a modality $X$, where $X\in \text{set of modalities }M=\{\text{images}, \text{words}\}$, $p({\vb{z}'_x}^L|\vb{z}_{M-\{x\}}^H)$ generates low-level events from high-level events of the modality $M - \{X\}$ and $p(\vb{x}'|{\vb{z}'_x}^L)$ regenerates the modality $X$. 
\begin{equation}
    \vb{z}'_x=p({\vb{z}'_x}^L|\vb{z}_{M-\{x\}}^H)~\text{and}~\vb{x}'=p(\vb{x}'|{\vb{z}'_x}^L)
\end{equation}

\subsection{Training Metrics}
\label{sec:approach:Learning Objective}
\subsubsection{Soft-Dynamic Time Warping (Soft-DTW)}
Given two trajectories $\vb{x}=(\vb{x}_1,\vb{x}_2,...\vb{x}_n)$ and $\vb{y}=(\vb{y}_1,\vb{y}_2,...\vb{y}_m)$, the $\text{soft-DTW}(\vb{x},\vb{y})$ (\cite{cuturi2017soft}) computes the discrepancy between $\vb{x}$ and $\vb{y}$ as,
\begin{equation}
    \text{soft-DTW}(\vb{x},\vb{y})= min^\gamma\{\langle A ,~\Delta (\vb{x}, \vb{y})\rangle, A \in ~\mathcal{A}_{n,m} \}
\end{equation}
where $A \in \mathcal{A}_{n,m}$ is the alignment matrix, $\Delta (\vb{x}, \vb{y})=[\delta(\vb{x}_i,\vb{y}_i)]_{ij} \in \mathcal{R}^{n\times{}m}$ and $\delta$ being the cost function. $min^{\gamma}$ operator is then computed as, 
\begin{equation}
    min^\gamma\{\vb{a}_1,\cdots,\vb{a}_n\}=
    \begin{cases}
        min_{~i\leq n}~\vb{a}_i,&\gamma=0, \\
        -\gamma\log\sum_{i=1}^{n}e^{-\vb{a}_i/\gamma},&\gamma>0.
    \end{cases}
\end{equation}
For our experiments, we use $L_{2}$ distance as $\delta$ and $\gamma=1$.

\begin{figure*}[t]
    \centering
    \includegraphics[width = 0.8\textwidth]{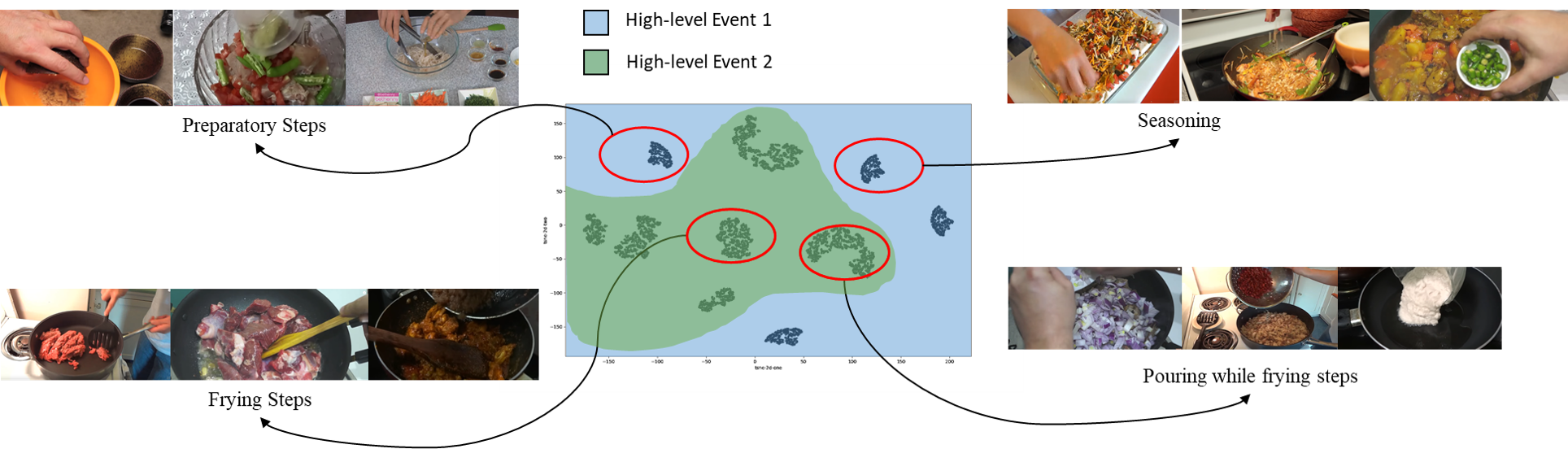}
    \caption{t-SNE of low-level events and their corresponding high-level mappings discovered by SHERLock on the YouCook2 dataset. We obtain clusters of low-level events such as frying, pouring while frying, seasoning etc. We also obtain two high-level events that correspond to events that require heating and those that do not.}
    \label{fig:tsne}
\end{figure*}
\subsubsection{Learning Objective}
We emphasize that we do not require supervision for hierarchical temporal segmentation, i.e., we do not require annotations which demarcate the beginning and ending of a event, both in language and in the space of frame's timestamps. 
Our approach uses several loss terms between network outputs to achieve our objective.
\begin{equation}
    % \small
    \begin{split}
        \mathcal{L}_{dyn}=\text{soft-DTW}(\vb{Z}_w^L,{\vb{Z}'}^{L}_w) + \text{soft-DTW}(\vb{Z}^L_s,{\vb{Z}'}^{L}_s) + \\
        \text{soft-DTW}(\vb{S},\vb{S}') +\text{soft-DTW}(\vb{W},\vb{W}') \\
    + \text{soft-DTW}(\vb{Z}^H_s,\vb{Z}^H_w)
    \end{split}
\end{equation}
% We use the Negative Log-Likelihood (NLL) loss ($\mathcal{L}_{nll}$) between the re-generated comment vectors $\vb{W}'$ and the BERT vectors $\vb{W}$ as follows,
\begin{equation}
    \begin{split}
    \mathcal{L}_{static}=L_{2}(\vb{Z}^H_s,\vb{Z}^H_w) +  L_{2}(\vb{Z'}^L_s,\vb{Z'}^L_w)
    \end{split}
\end{equation}
We then define our total loss as,~$\mathcal{L}_{total}= \mathcal{L}_{dyn}+  {\beta}*\mathcal{L}_{static}$. We posit that this loss function provides the inductive bias necessary for learning the event latent space. The term $\text{soft-DTW}(\vb{S},\vb{S}')$ ensures reconstruction of demonstration frames from the textual events, while $\text{soft-DTW}(\vb{W},\vb{W'})$ ensures the generation of textual description from visual events. $L_{2}(\vb{Z}^H_s,\vb{Z}^H_w)$ and $L_{2}(\vb{Z'}^L_s,\vb{Z'}^L_w)$ aligns the textual and visual event spaces.

\subsection{Evaluation Metrics}
\label{sec:approach:eval_metrics}
The ground-truth events in the dataset and the events generated by SHERLock may differ in number, duration, and start-time. To evaluate the efficacy of SHERLock in generating events that align with the human-annotated events in our dataset, it is imperative that we utilize a metric that measures the overlap between generated events and ground truths and also accounts for this possible temporal mismatch. %To this end, we introduce \emph{Time-Warp IoU}.

Consider the search series $\vb{X}=(\vb{x}_1,\vb{x}_2,\vb{x}_3...\vb{x}_M)$ and target series $\vb{T}=(\vb{t}_1,\vb{t}_2,\vb{t}_3...\vb{t}_N)$ where $\vb{X}$ corresponds to the end-of-event time stamp for each event as generated by SHERLock for a single long-horizon demonstration trajectory. Thus, the $i^{th}$ event abstracted from SHERLock starts at time $\vb{x}_{i-1}$ and end at time $\vb{x}_{i}$. Similarly, $T$ corresponds to the end-of-event time stamp for each ground-truth event in the demonstration trajectory, where the $j^{th}$ ground truth event starts at time $\vb{t}_{j-1}$ and ends at time $\vb{t}_{j}$. Note that both $\vb{x}_0$ and $\vb{t}_0$ are equal to zero i.e. we measure time starting at zero for all demonstration trajectories.

To meaningfully compute the \textbf{intersection over union (IoU)} between ground truth and outputs from SHERLock, we first need to align the two representations using dynamic time warping (DTW; \cite{berndt1994using}). This implies calculating $\Delta (\vb{X}, \vb{T})$, solving the following DTW optimization problem (\cite{berndt1994using}), $\Delta (\vb{X}, \vb{T}) = \min_{\vb{P} \in \mathcal{P}} \sum_{\vb{m}, \vb{n} \in \vb{P}} \delta(\vb{x}_m,\vb{t}_n)$,

where the $\vb{X}$ and $\vb{T}$ correspond to the search and target series respectively and $\delta$ corresponds to a distance metric (in our case the $L_2$ norm), measuring time mismatch.

$\Delta (\vb{X},\vb{T})$ therefore corresponds to the trajectory discrepancy measure defined as the matching cost for the optimal matching path $\vb{P}$ among all possible valid matching paths $\mathcal{P}$ (i.e., paths satisfying monotonicity, continuity, and boundary conditions). From this optimal trajectory we can also obtain the warping function $W$ such that $W(\vb{x}_i) = \vb{t}_j$, i.e. we find the optimal mapping between the $i^{th}$ event ending at time $\vb{x}_i$ and the $j^{th}$ event ending at time = $\vb{t}_j$. The resulting Intersection over Union for a single long-horizon trajectory, Time-warped IoU ($\text{TW-IoU}$), is:  

\begin{equation}
\small
% \begin{split}
    % TWIoU = 
    \sum_{t_i} \frac{\sum_{x_j:W(x_j)=t_i} \min(t_i, x_j) - \max(t_{i-1},x_{j-1})}
            {\splitfrac{\max_{x_j:W(x_j)=t_i} \{\max(t_i, x_j)\}}  {-\min_{x_j:W(x_j)=t_i}\{\min(t_{i-1},x_{j-1})\}}}
% \end{split}
\end{equation}
More details are presented in Section \ref{twiou-d} of Supplementary Material. 
\subsection{Alignment during Inference}
We calculate the DTW path \cite{berndt1994using} ($\gamma = 0$ case in eqn. (9)) between a decoded sequence and a ground truth video to obtain the optimal alignment between ground truth video frames and predicted video frames (high \& low-level). This alignment is subsequently used during the calculation of the TW-IoU scores.

% \aboverulesep=0ex
% \belowrulesep=0ex
% \captionsetup{belowskip=-4pt,aboveskip=4pt}
\begin{table}[t]
% \begin{table*}[thb!]
% \begin{wraptable}{r}{8cm}
% \vspace{-4pt}
    % \centering
    \begin{minipage}{.99\linewidth}
        \centering
        \small
        \begin{tabular}{|c|c|}
            % \hline
            \toprule
            \textbf{Method} & \textbf{TW-IoU} \\
            % \hline
            \bottomrule
            % Random  & \multicolumn{2}{c|}{28.2} \\
            % EQUALDIV  & \multicolumn{2}{c|}{31.7} \\
            Non-Hierarchical & \\ \cite{Shankar2020Discovering} & $14.47 \pm 1.10$ \\\hline
            Non-Hierarchical w/ comment & $14.84 \pm 1.08$ \\\hline
            GRU Change Point Prediction & $22.85 \pm 0.74$ \\\hline
            Clustering (ResNet32) &\\ (\cite{he2016deep}) & $31.22 \pm 0.05$ \\\hline
            Clustering (HowTo100M) &\\ (\cite{miech2019howto100m}) & $32.17 \pm 0.04$ \\\hline
             
            \hline
            SHERLock-GRU &\\ w/o comment (ours) & $35.99 \pm 1.13$\\\hline
            SHERLock &\\ w/o comment (ours) & $39.45 \pm 1.25$\\\hline
            \textbf{SHERLock} &\\ w/ comment (ours) & $\textbf{47.44} \pm \textbf{1.64}$\\\hline
            \hline
            GRU-Supervised &\\ Segment Prediction & \textbf{$53.12 \pm 1.09$} \\
            \hline
        \end{tabular}
        \caption{TW-IoU scores for single level events predicted by the baselines along with the TW-IoU scores for the high-level events abstracted by our proposed technique SHERLock.}
        \label{tab:baseline}
        \end{minipage}\hfill
        \end{table}
\begin{table}[h]
        \begin{minipage}{.99\linewidth}
        \centering
        \small
        \begin{tabular}{|c|c|}
            \toprule
          \textbf{Ablation Variants} & \textbf{TW-IoU} \\
            \toprule
            SHERLock & \\ w/o comment w/o L2 loss & 38.46\\\hline
            SHERLock & \\ w/o comment w/o cross-decoding & 37.67\\\hline
            SHERLock \\ Single-level Decoding  & 20.33\\\hline
            SHERLock & \\ w/o comment w/o low-align loss & 18.99\\\hline
            SHERLock & \\ Three-level Hierarchy & 37.61\\\hline
            SHERLock & \\ w/o comment (200 Frames) & 39.45\\\hline 
            SHERLock & \\ w/o comment (64 Frames) & 33.41\\\hline 
            SHERLock & \\ w/o comment (32 Frames) & 12.79\\
            \hline
        \end{tabular}
        \caption{TW-IoU scores for ablation experiments. Note that the reported TW-IoU scores are calculated with reference to high-level annotations available in the dataset (low-level annotations are unavailable). See Figure 8 of Supplementary Material for model architectures.}
        \label{tab:ablation}
        \end{minipage}%
\end{table}
\begin{table}[h]
    \centering
        \centering
        \small
        \begin{tabular}{|c|c|}
            % \hline
            \toprule
            \textbf{Method} & \textbf{TW-IoU} \\
            % \hline
            \bottomrule
            % Random  & \multicolumn{2}{c|}{28.2} \\
            % EQUALDIV  & \multicolumn{2}{c|}{31.7} \\
            Non Hierarchy (\cite{Shankar2020Discovering}) & $4.47 \pm 1.12$  \\
            \hline
            SHERLock w/o comment (ours) & $40.69 \pm 1.66$ \\
            \hline
            SHERLock w/ comment (ours) & $\textbf{52.66} \pm \textbf{1.72}$ \\
            \hline
        \end{tabular}
        \caption{TW-IoU scores on the TutorialVQA dataset}
        \label{tab:task}
    \end{table}
\section{Experiments}

\label{sec:experiments}
% We use the following datasets for our experimentation: 
\textbf{Datasets:} \textbf{YouCook2}(\cite{ZhXuCoCVPR18}) dataset comprises of instructional videos for 89 unique food recipes. 
\textbf{Recommending Chess Openings}\footnote{https://www.kaggle.com/residentmario/recommending-chess-openings} dataset consists of opening moves in the game of Chess. 
\textbf{TutorialVQA} (\cite{colas2019tutorialvqa}) consists of 76 tutorial videos pertaining to an image editing software. For dataset details, see Sec \ref{sec:Datasets} in the Appendix. For \textbf{implementation details}, see Sec \ref{sec:imple} in the Appendix.
% and show that our architecture is able to identify the correct label, even for chess games in 
% , into openings and their corresponding variation in an unsupervised manner - a task that requires significant knowledge of chess. 

% \paragraph{Evaluation Metrics}
% % https://arxiv.org/pdf/1901.06829.pdf
% IOU and Alignment score

% We perform all experiments on Ubuntu 14.04 machine with 16 X Nvidia A100 GPUs and Intel(R) Xeon(R) CPU. It additionally had 96 CPU cores and 1.31TB of RAM.

\begin{figure}[t]
    \centering
    \includegraphics[width = 0.9\linewidth]{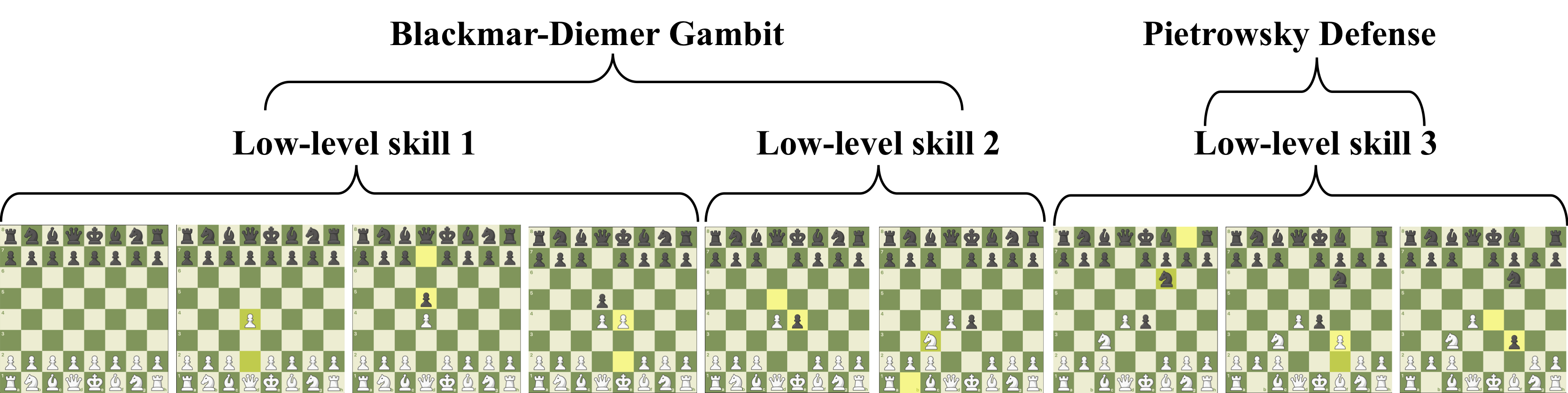}
    \caption{Hierarchy of events discovered by SHERLock using openings data in Chess.
    At a high level, SHERLock correctly identifies events corresponding to the Blackmar-Diemer Gambit and the Pietrowsky Defense.
    At a low level, it identifies events such as ``\textit{d4 d5..} and \textit{e4 d3 ..}'' that are used across several openings. More encouraging results are presented in the Appendix \ref{chess_results}}
    \label{fig:chess_hie}
\end{figure}

\subsection{Visualizing Hierarchy}
\label{sec:experiments:visualizing_hierarchy}
\begin{figure*}[h]
    \centering
    \includegraphics[width = 0.8\textwidth]{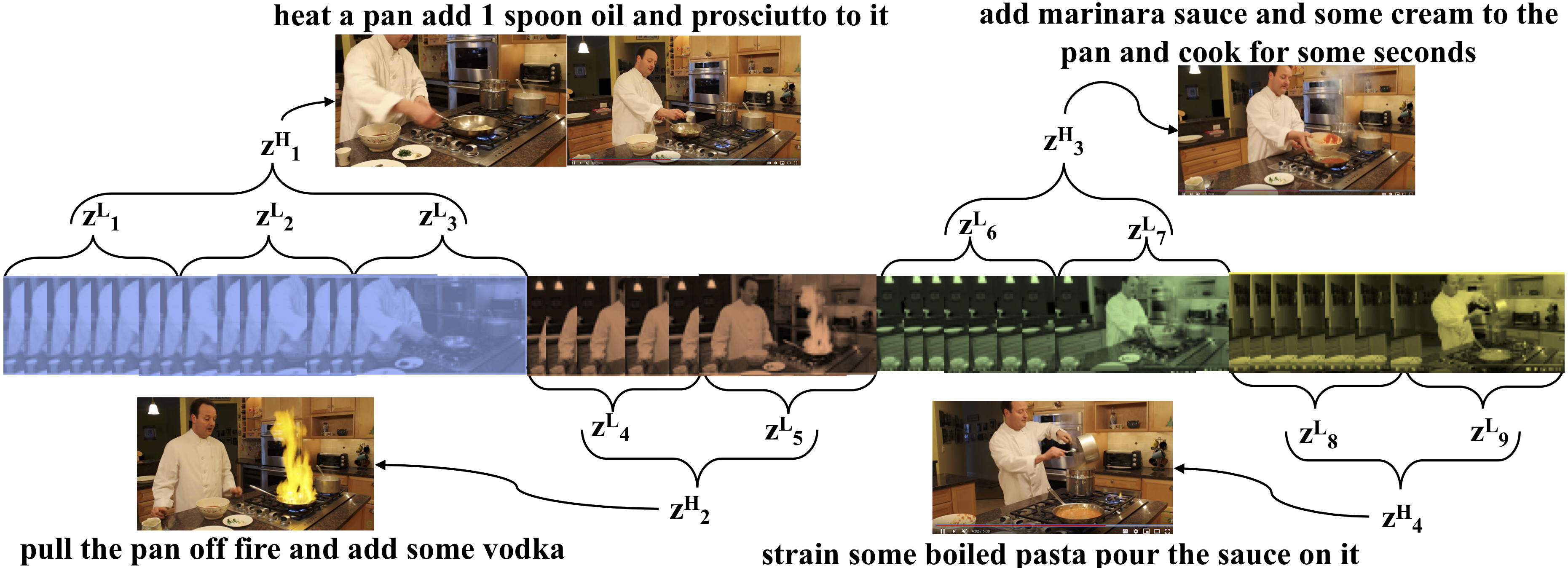}
    \caption{Example Hierarchy of events discovered by SHERLock on the YouCook2 dataset}
    \label{fig:cook_hie}
\end{figure*}

Here we analyze whether the discovered events are human interpretable i.e. \textit{are the temporal clusters within a single demonstration semantically meaningful?}.
We find that SHERLock abstracts several useful human interpretable events without any supervision.
See Figure~\ref{fig:chess_hie} and ~\ref{fig:cook_hie} for results.
% The abstracted high-level events align well with the ground truth event labels provided in YouCook2.
% We also additionally find that our model is able to split each high-level event into meaningful low-level events.
For instance, in a pasta-making demonstration in YouCook2, a single event corresponding to the description ``\textit{heat a pan add 1 spoon oil and prosciutto to it}'', is divided into low level events corresponding to ``\textit{heat pan}'', ``\textit{add oil}'' and ``\textit{prosciutto}''. Also in Figure~\ref{fig:phototshop_hie}, a single high-level event corresponding to ``\textit{editing image text}'' is divided into low level events like ``\textit{changing text color}'', ``\textit{text font}'', ``\textit{typekit font}'', etc.
Note that no explicit event time labels were provided to SHERLock, indicating that our model can abstract such coherent sub-sequences, thus taking the first step towards video understanding.
Figure \ref{fig:tsne} shows the t-SNE \cite{maaten2008visualizing} for the low-level event representations abstracted by SHERLock. The low-level events aggregate into clusters corresponding to frying, pouring while heating, seasoning.
We also visualize the events abstracted by SHERLock when trained on chess opening data. The events learnt here also produce coherent, human-interpretable results. See Fig- \ref{fig:all-architectures-abl} onwards in Supplementary Material for more visualizations.

\subsection{Comparison with Baselines}
We evaluate the performance of SHERLock quantitatively on YouCook2 and TutorialVQA and quantify its ability to generate coherent events that align with the human annotated ground truths using the TW-IoU metric. We compare our approach with 6 baselines.
\leavevmode\newline \textbf{GRU Time Stamp Prediction}: A supervised baseline comprising of a GRU-based encoder \cite{cho-etal-2014-learning} that sequentially processes the ResNet features corresponding to frames in a video followed by a decoder GRU \cite{bahdanau2014neural} that attends to encoder outputs and is trained to sequentially predict end-of-event timestamps of each meaningful segment (variable in number) in the video.
\leavevmode\newline \textbf{Non-Hierarchical w/o comment}: We implement the \cite{Shankar2020Discovering} approach (SOTA in unsupervised skill learning w/o environment) which takes as input a sequence of video frames and discovers a single level of events without any hierarchy. 
\leavevmode\newline \textbf{Non-Hierarchical w/ comment}: A modified multi-modal version of Non-Hierarchical where frames and {words} are utilized to form a non-hierarchical latent event representation. This baseline ascertains the effect of both hierarchical and multi-modal learning on the representations obtained. See Fig \ref{fig:all-architectures-abl} in Supplementary Material for architectural details.
\leavevmode\newline \textbf{Clustering - ResNet32 Embeddings}: Given an input sequence of frames, we define the weight function based on their temporal position in the sequence and also the $L_2$ distance between the frame embeddings. Then we use standard K-means algorithm (we find best K=4) to cluster the frames based on the weighting function defined and use the clusters formed to predict the temporal boundaries.
\leavevmode\newline \textbf{Clustering - HowTo100M Embeddings}: We utilize the pre-trained embeddings from the supervised action recognition dataset and method \cite{miech2019howto100m} and apply a K-means (we find best K=4) clustering on them. 
\newline \textbf{GRU Supervised Segment Prediction}: Instead of predicting end time stamps of each segment (as in GRU Time Stamp Prediction), the decoder is trained to predict/assign identical ids to frames which are part of the same segment. Further, the model’s decoder is trained to assign different ids to frames part of different segments while frames not part of any meaningful segment in the ground truth are trained to have a default null id - 0. 
%During inference, continuous subsequence of frames predicted to be having same id are considered as part of one segment and different predicted segments are extracted accordingly (frames predicted to be having null ids are ignored).

% (1) \textbf{Random} baseline predicts segment randomly on the basis of uniformly sampled timestamps for a given video on the basis of its duration.
% (2) \textbf{EQUALDIV} consists of dividing the video into conceptual segments of equal duration
Table \ref{tab:baseline} summarises and compares the TW-IoU computed between ground truth time stamp annotations and predicted/discovered segments. SHERLock achieves the highest TW-IoU when compared with all other unsupervised baselines. We find that SHERLock discovers events that align better with the ground truth events (\textbf{SHERLock performs $\sim23\%$ better}) compared to Non-Hierarchical \cite{Shankar2020Discovering} performing at par with the supervised baselines.

% 

% \subsection{Label Prediction Task for Chess Openings}
% We evaluate SHERLock on the task of label (name of opening + variation) prediction using the hierarchical event (in this case, strategy) representations discovered on the Chess Openings dataset. The dataset comprises of a label for each sequence of moves in an opening with $300$ distinct labels for each opening and variation. We train a simple linear classifier which uses the embedding generated by SHERLock (Figure~\ref{fig:net_architecture}) to predict the label and find that it achieves \textbf{78.2\%} accuracy of prediction. Thus, the representations abstracted by SHERLock contain temporal information that aligns with the human understanding of chess strategies. SHERLock represents openings and variations as hierarchical events and is able associate them with the correct label.

% \subsubsection{Language}

% \subsubsection{Non-Hierarchical Skills}
% Language, sampling rate, Comparison with Non-Hierarchical skills
% \subsection{Ordering Task}
\subsection{Ablation Experiments}
\label{sec:ablations}
\leavevmode\newline\textbf{Effect of Sampling Rate on the quality of hierarchy}
For YouCook2, we cap the length of a frame sequence to $200$ frames (down-sampled from the original frames provided in the dataset due to memory constraints). Subsequently, we analyze the trade-off between sequence length and performance. This provides an insight into granularity of information required to discover naturalistic hierarchies. Interestingly, we don't observe a linear drop in performance with a reduction in the number of frames (refer Table \ref{tab:ablation}).

\leavevmode\newline\textbf{Effect of Guidance through Commentary}
We study the effect that language has on event discovery by comparing SHERLock without comment (Fig\ref{fig:cook_hie_lang} in Supplementary), which discovers event hierarchy using just frames and SHERLock which additionally uses word embeddings as a guide (as in Figure~\ref{fig:net_architecture}). \textbf{Language improves the TW-IoU by $\sim10\%$}, indicating that using commentary enables SHERLock to detect more precisely the boundaries of segments corresponding to various events in a trajectory. Further, we find (Fig-\ref{fig:all-architectures-abl} in supplementary) that the implicitly hierarchical nature of the language provides inductive bias to the model to learn a more natural hierarchy of events.

\leavevmode\newline\textbf{Number of Levels in Hierarchy}
We explore the effect of a third level of hierarchy, through additional transformers during the encoding and decoding phase. Thus, our architecture generates 16 low-level, 8 mid-level and 4 high level events. We find that this third level of event \textbf{provides only a marginal improvement} over the TW-IoU scores which we report in table \ref{tab:ablation}. Additionally, we find that this increases the GPU memory requirements during training due to the increased number of model parameters in memory along with the additional losses (calculating Soft-DTW losses means solving a dynamic programming problem).
% \subsection{Model Complexity}

\leavevmode\newline\textbf{Model Complexity:}
SHERLock uses the Transformer architecture for modeling $\Phi$ and $p()$ which makes the model a bit heavy to train. So, we experiment by replacing all the transformer modules with simpler GRU modules keeping same number of layers (See SHERLock-GRU in Table~\ref{tab:baseline}). We observe that there is \textbf{not much difference} in performance ($\sim$ 3.5\%). Also it still outperforms all other unsupervised baselines. This indicates that the attention mechanism in Transformers does help us learn better representations but most of the gain can be attributed to the model architecture.

\label{sec:three_lev}
\leavevmode\newline\textbf{Components and Losses:}
We perform ablation experiments to ascertain the need for each of the modules and losses used in SHERLock. We remove the $\text{soft-DTW}(\vb{Z}^L_s,{\vb{Z}'_s}^{L})$ loss from our SHERLock to highlight its importance in maintaining the fidelity of the reconstruction scheme. This loss guides the alignment between the encoded low-level events ($\vb{Z}^L_s$) and the reconstructed low-level events (${\vb{Z}'_s}^{L}$). We find that \textbf{removing this loss reduces the TW-IoU scores drastically} (see \textit{SHERLock w/o comment w/o low-align loss} in Table~\ref{tab:ablation}). 

We also evaluate a simplified version of the SHERLock w/o commentary model, where we remove the ${\vb{Z}'}^L={\vb{z}'_{0:7}}^L~\sim p({\vb{z}'}^L|\vb{z}^H)$ modules and re-generate the word and visual sequence embeddings from the high-level events as $\vb{X}'=\vb{x}'_{0:T}~\sim p(\vb{x}'|\vb{z}^H)$. We see this results in the drop of TW-IoU (Table~\ref{tab:ablation}), thus confirming our need for the step-wise encoding-and-decoding scheme used. We call this the \textit{SHERLock Single-level Decoding} baseline the diagram for which is included in the Supplementary Material (Fig-\ref{fig:all-architectures-abl}).

\section{Conclusion}
\label{sec:conclusion}
In this paper, we provide a self-supervised method (SHERLock) capable of hierarchical and multi-modal learning. It can discover events and organize them in a meaningful hierarchy using only demonstration data from chess openings, tutorials, and cooking. 
We also show that this discovered hierarchy of events helps predict textual labels and temporal event segmentations for the associated demonstrations. 
% However, even though SHERLock outperforms other baselines we believe there is a lot of scope for improvement as per the TW-IoU metric.

One limitation is that we can't use longer video sequences for training since computing soft-DTW requires solving a DP problem in quadratic space. Also sometimes specific nouns like \textit{"lobster"} are replaced with more commonly appearing nouns like \textit{"patty"}, which is due to the fact that grounding of nouns using a few images is very difficult. This could be an interesting direction for future work.
Also, we would explore curriculum learning where the discovered event hierarchy by SHERLock is used to generate curricula where lower-level events would be taught first followed by higher-level events and also be used for option discovery and training in reinforcement learning.


%\section*{Acknowledgment}




\bibliographystyle{IEEEtran}
\bibliography{bibliography}


% \end{thebibliography}

\clearpage

\appendix
% \section{\textit{Appendix for}\\
% SHERLock: Self-supervised Hierarchical Event Representation Learning
% }

% \section{Additional Experiments \& Details}

\subsection{Implementation Details}
\label{sec:imple}
We down-sample video frames per trajectory to 200 frames and encode each frame with ResNet-32 (pretrained on MSCOCO dataset) (\cite{he2016deep}) to a $512\times{}1$ dimension embedding. Comments are encoded using BERT-base pre-trained embeddings with a $768$ hidden dimension. Each of the $p(\vb{z}|\vb{w})$, $p(\vb{z}|\vb{s})$, $p(\vb{s}|\vb{z})$, $q(\vb{w}|\vb{z})$ modules consist of the Transformer (\cite{vaswani2017attention}) Encoder with 8 hidden layers and 8-Head Attention which takes as input, a positionally-encoded sequence and outputs attention weights. It is then passed through a Transformer Decoder with 8 hidden layers to generate latent variables having dimension $event length\times{768}$.
% and the encoder-decoder cross-attention with masking. which finally outputs the events which are latent variables having dimension $event length\times{768}$. 
We use a 1-layer GRU (\cite{chung2014empirical}) for each of the $p(\vb{z}|\vb{z})$ modules to generate low-level events from  high-level events. 
%We provide additional details about the parameters and the hardware used for training in Appendix~\ref{appendix:experimental_details}.

We keep the maximum number of events discovered at low-level to 16 and high-level at 4.
These assumptions are based on the YouCook2 dataset statistics where the minimum number of segments were 5 and the maximum as 16.
We train the network with Adam optimizer for 100 epochs with $lr=1e-5$, $\alpha = 1$ and $\beta = 1$ for all our experiments along with a batch-size of 128. We use 16x Nvidia A100 GPUs to train SHERLock. Results are reported based on 5 runs with random seeds. It takes $\sim$ 48 hours to reproduce the reported results.
\subsection{Dataset Details}
\label{sec:Datasets}
\textbf{YouCook2} consists of recipes ($\sim$22 videos per recipe) containing labels that separate the long horizon trajectories of demonstrations into events - with explicit time stamps for the beginning and end of each event along with the associated commentary. 
% We show that our architecture identifies events in an unsupervised manner without access to the ground truth event labels. It also discovers lower-level concepts that divide the ground truth events into simpler semantically meaningful subsequences. We also study the ability of the architecture to produce coherent textual descriptions. 
It contains 1,333 videos for training and 457 videos for testing. The average number of segments per video is 7.7 and the average duration of the video is 5.27 minutes.
We provide all the modified system diagrams for all the ablation experiments mentioned in Section 3.4 for clarity.
\newline\textbf{Chess Opening: }An Opening in Chess is a fixed sequence of moves which when performed leads to a final board state putting the player in a strategic position in the game. Commonly used chess openings are each labeled with a name (Figure~\ref{fig:chess_hie} shows examples. See Figures \ref{fig:French}, \ref{fig:Sicilian1}, \ref{fig:Sicilian3} in Supplementary Material). The dataset contains 20,058 openings with each containing a sequence of chess moves and it's corresponding opening and variation labels. The train-test split used for our experiments is 80-20.
\newline\textbf{TutorialVQA: }All videos include spoken instructions which are transcribed and manually segmented into total 408 segments. The average transcript length is 48. 

\subsection{Visual Ordering Task}
Both SHERLock and Non-Hierarchical discover event representations in an unsupervised manner. To show the effectiveness of the representations discovered we use them to perform a Visual Ordering Task. The task is that given a sequence of frames as input, our trained model (with frozen weights) should discover the high-level events associated with those frames and can be used to predict whether or not the given sequence of frames are in correct/meaningful order (binary classification). We use a simple 1-layer GRU to do so. To create the training data for this we take examples from the YouCook2 (\cite{ZhXuCoCVPR18}) dataset and randomly shuffle the sequence of frames creating 10 negative examples for each positive sample in the dataset. As we can see that there is a significant gain of \textbf{12\%} in the F1 score and \textbf{2\%} in accuracy using our model SHERLock over the Non-Hierarchical baseline.

\begin{table}[th]
    \centering
        \centering
        \small
        \begin{tabular}{|c|c|c|}
            % \hline
            \toprule
            \textbf{Method} & \textbf{Accuracy} & \textbf{F1} \\
            % \hline
            \bottomrule
            % Random  & \multicolumn{2}{c|}{28.2} \\
            % EQUALDIV  & \multicolumn{2}{c|}{31.7} \\
            Non-Hierarchical (\cite{Shankar2020Discovering}) & 88.9 & 2.13 \\
            SHERLock w/o comment (ours) & \textbf{90.9} & \textbf{14.9}\\
            \hline
        \end{tabular}
        \caption{Visual Ordering Task}
        \label{tab:vorder-task}
    \end{table}

\subsection{Details about TW-IoU}
\label{twiou-d}
TW-IoU takes into account the original temporal boundaries. For every interval in the generated event segments, the time-warped alignment function finds the optimal mapping for that interval to one of ground-truth intervals provided in the dataset.  
For example, suppose that the ground-truth segment consists of 2 events: $[0,a]$ and $[a,a+b]$ i.e., the first event starts at time 0 ends at time $a$, and second starts at $a$ and ends at $a+b$). Consider two different sets of aligned predictions : 
\begin{enumerate}
    \item one predicted event - $[0,c+d]$
    \item two predicted events $[c,c+d]$ and $[c,c+d]$
\end{enumerate}

Without any loss of generality, we assume $c > a$ and $d > b$ and all a,b,c,d are positive real numbers. For the above:
\begin{enumerate}
    \item TW-IoU =  $\frac{(a+b)}{(c+d)}$
    \item TW-IoU =  $\frac{(a)}{(c)}+\frac{(b)}{(d)}$
\end{enumerate}

Now it’s trivial to show that TW-IoU is greater for (2) than (1) under the above assumptions. We would prefer (2) over (1) as there are indeed 2 ground-truth events in the search series. Thus, TW-IoU performs as expected. This is what we would expect to happen since (2) is a better prediction as it breaks down the original high-level input to find 2 low-level events whereas (1) doesn’t break it down hierarchically. Thus we would want our proposed metric TW-IoU to perform better in case (2) which is exactly what will happen. We hope this clears our the motivation behind this. Also adding to this note that the alignment function in TW-IoU works sequentially, i.e. it won’t align any interval with current ground-truth being processed until all previous intervals have been aligned.

% Below are the architectures for some of the ablation variants as mentioned in Table 2 in the paper and few more qualitative results on the \textit{Recommending Chess Openings} dataset: 
\FloatBarrier
\subsection{Ablation Architectures \& Results}
\begin{figure*}[t]
    \centering
    \includegraphics[width = \textwidth]{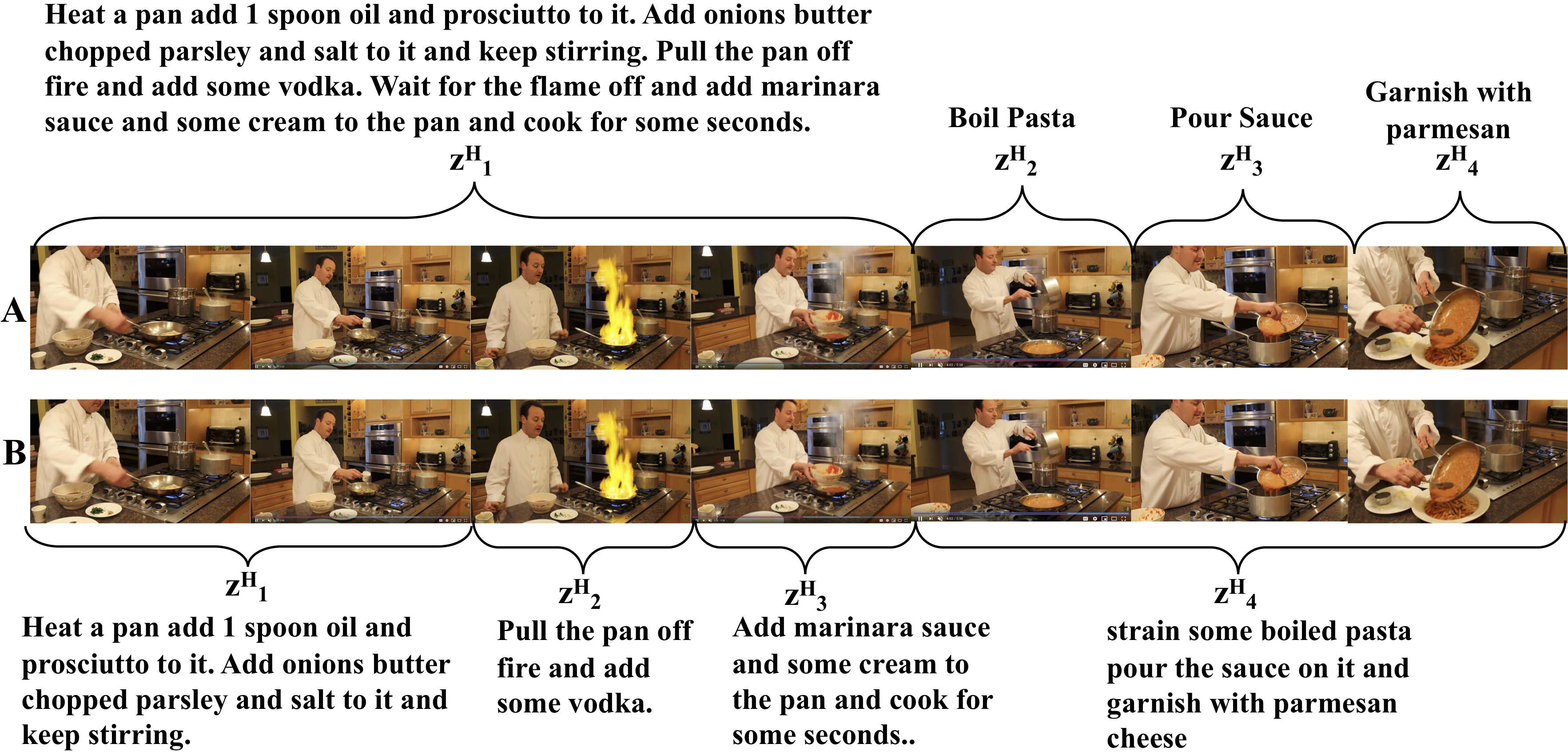}
    \caption{In this figure, we show that using commentary as guide during training, SHERLock learns to better combine low-level events to form high-level events which are better aligned towards our ground-truth annotations. \textbf{A} refers to the event segments discovered w/o comments and \textbf{B} refers to the one with comment.}
    \label{fig:cook_hie_lang}
\end{figure*}

\begin{figure*}[htb!]
    \centering
    \includegraphics[width =  0.75\linewidth]{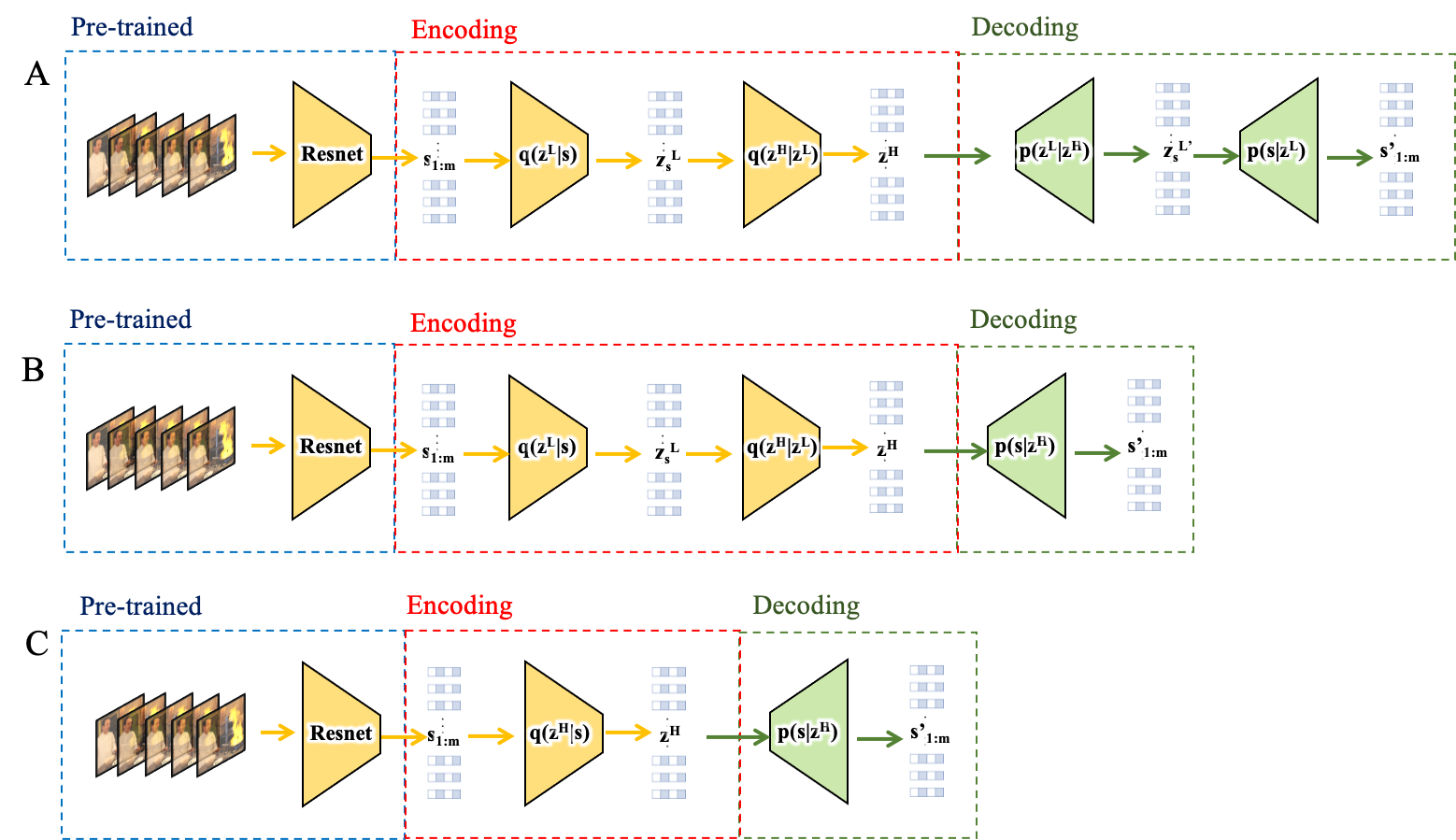}
    \caption{Three model variants are shown here. \textbf{(A)} refers to SHERLock w/o comment. \textbf{(B)} refers to the SHERLock Single-level Decoding variant where we directly predict the sequence of frames from the high-level event. \textbf{(C)} refers to Non-Hierarchial (Shankar et al.) baseline without comment.}
    \label{fig:all-architectures-abl}
\end{figure*}

See Figure~\ref{fig:cook_hie_lang} to see the effect of adding commentary as guide during training as mentioned in ~\ref{sec:ablations}. Also see Figure~\ref{fig:all-architectures-abl} for more details on ablation experiments.

\subsection{More results on \textit{Recommending Chess Openings}} \label{chess_results}
\begin{figure*}[b]
    \centering\includegraphics[width =  0.85\textwidth]{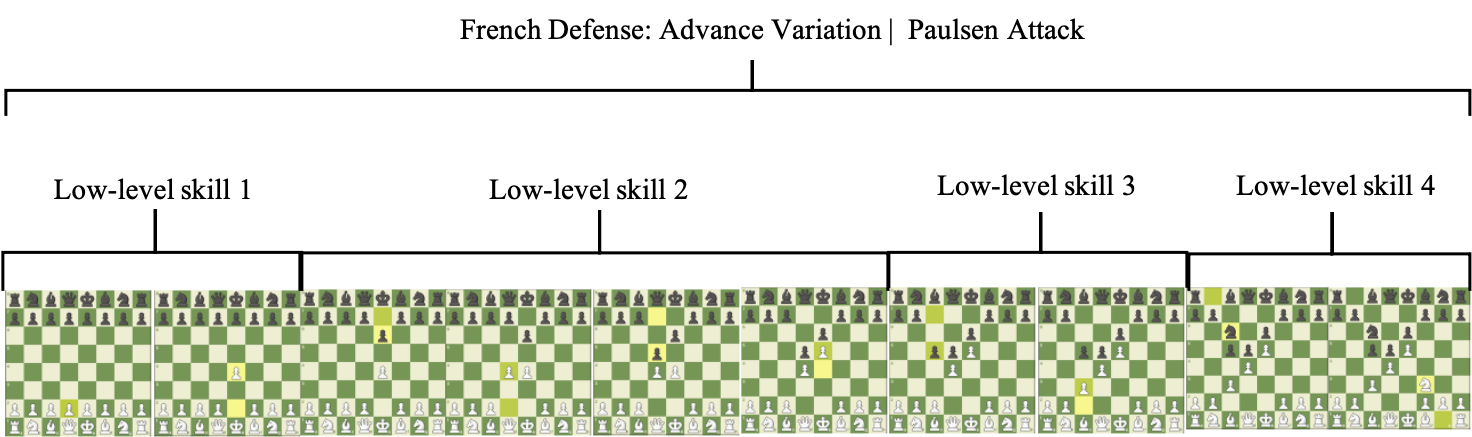}
    \caption{
    Skills discovered in the French defence.
    We see that low-level skill 1 corresponds to the kings pawn opening (e4).
    Further, low-level skill 2 correctly identifies the advance variation of the French defence.
    Finally, low-level skill 3 describes a common idea (c5) in the advance variation of the French defence.
    }
    \label{fig:French}
\end{figure*}

\begin{figure*}[t]
    \centering
    \includegraphics[width =  0.85\textwidth]{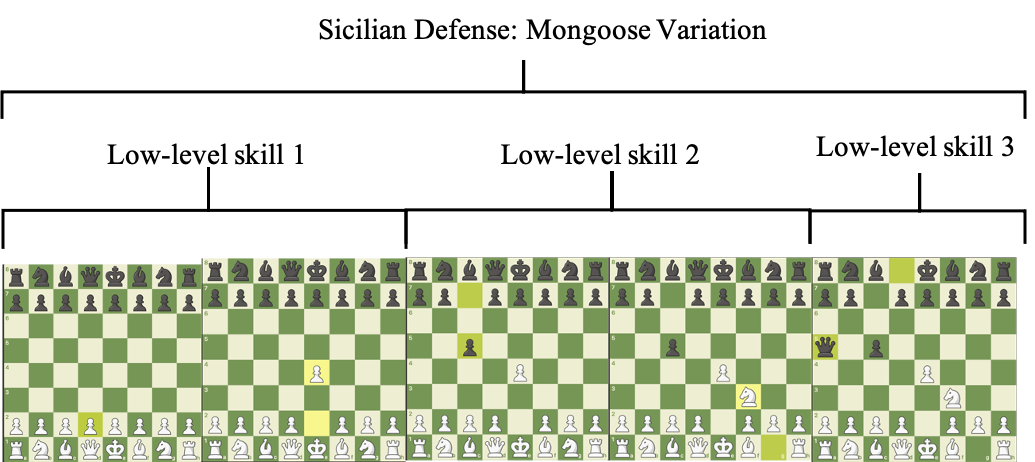}
    \caption{Skills discovered in the Sicilian opening.
    As before, we see that low-level skill 1 corresponds to the kings pawn opening (e4).
    Low-level skill 2 represents the Sicilian Defence (c5) and low-level skill 3 is the Mongoose variation of the Sicilian Defence. 
    Interestingly, our method automatically divides the Mongoose variation into the Sicilian Defence and the Qa5 idea.
    }
    \label{fig:Sicilian1}
\end{figure*}

Refer to Fig \ref{fig:French}-\ref{fig:Sicilian4}
\subsection{More Results on \textit{TutorialVQA}}
Refer to Fig \ref{fig:ps3}-\ref{fig:ps6}

\iffalse

\begin{figure*}[t]
    \centering
    \includegraphics[width =  0.75\textwidth]{Nimzo.png}
    \caption{Skills discovered in the Nimzowitsch Defense: Kennedy Variation :Linksspringer Variation opening.}
    \label{fig:Sicilian2}
\end{figure*}
\fi

\begin{figure*}[htb!]
    \centering
    \includegraphics[width =  0.75\textwidth]{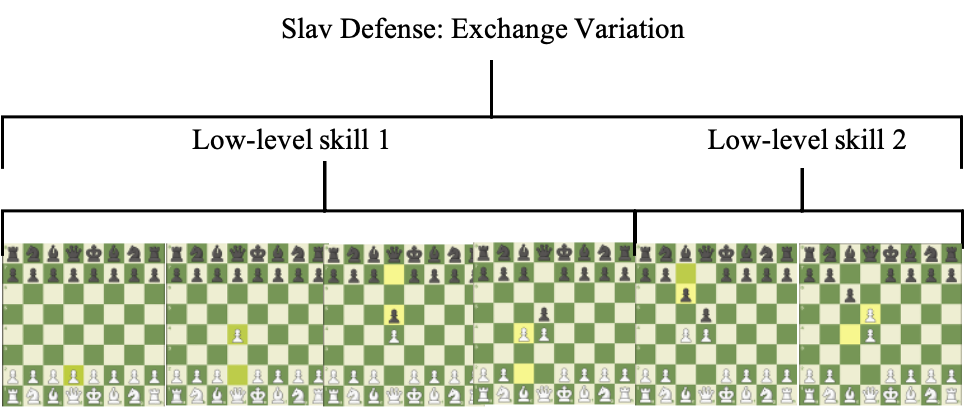}
    \caption{Skills discovered in the Slav Defense: Exchange Variation opening.
    We can see that low-level skill 1 captures the queen's gambit variation following d4.
    Further, low-level skill 2 captures the exchange variation of the Slav (dxe5).}
    \label{fig:Sicilian3}
\end{figure*}

\begin{figure*}[t]
    \centering
    \includegraphics[width =  \textwidth]{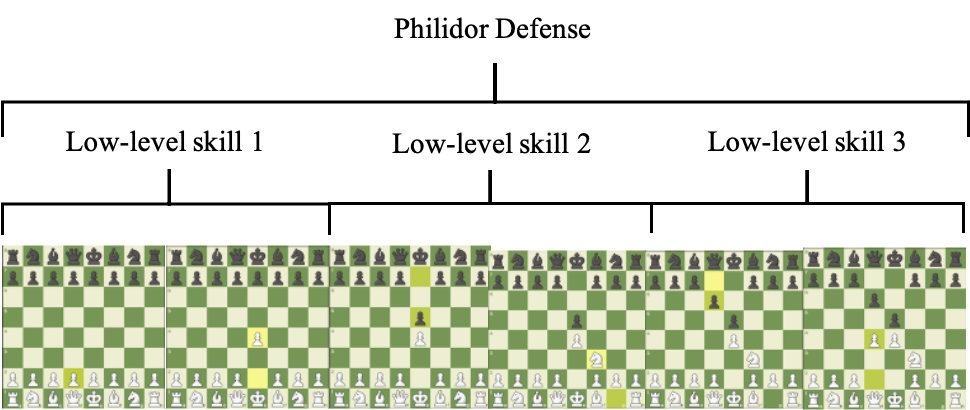}
    \caption{Skills discovered in the Philidor Defense opening.
    Low-level skill 1 describes the standard king's pawn opening (e4).
    Low-level skill 2 represents the Philidor Defence (Nf3 d6). }
    \label{fig:Sicilian4}
\end{figure*}
\FloatBarrier
% \subsection{More Results on \textit{TutorialVQA}}
\begin{figure*}[t]
    \centering
    \includegraphics[width = 1.0\textwidth]{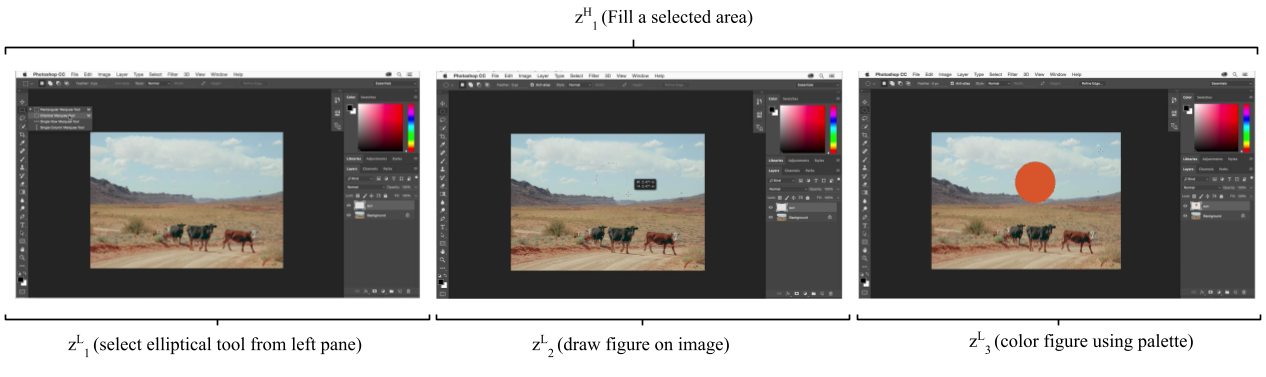}
    \caption{
 In this the high-level event \textit{Fill a selected area} is broken down into low-level events like \textit{select elliptical tool from left pane}, \textit{draw figure on image} and \textit{color figure using palette}.
    }
    \label{fig:ps3}
\end{figure*}
\begin{figure*}[t]
    \centering
    \includegraphics[width = 1.0\textwidth]{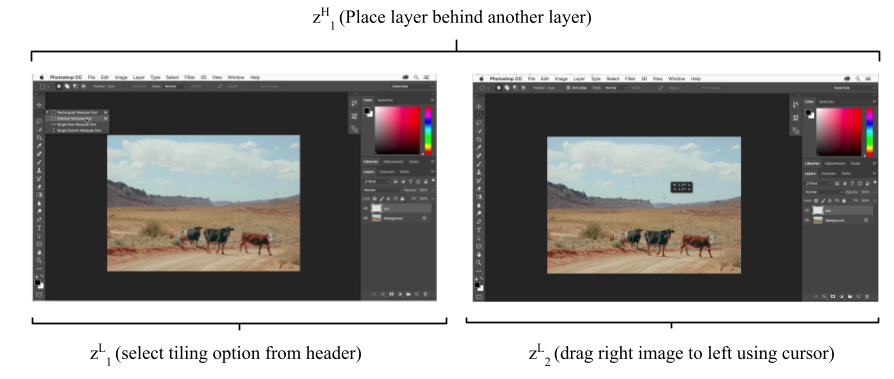}
    \caption{
 In this the high-level event \textit{Place layer behind another layer} is broken down into low-level events like \textit{select tiling option from header} and \textit{drag right image to left using cursor}.
    }
    \label{fig:ps4}
\end{figure*}
\begin{figure*}[t]
    \centering
    \includegraphics[width = 1.0\textwidth]{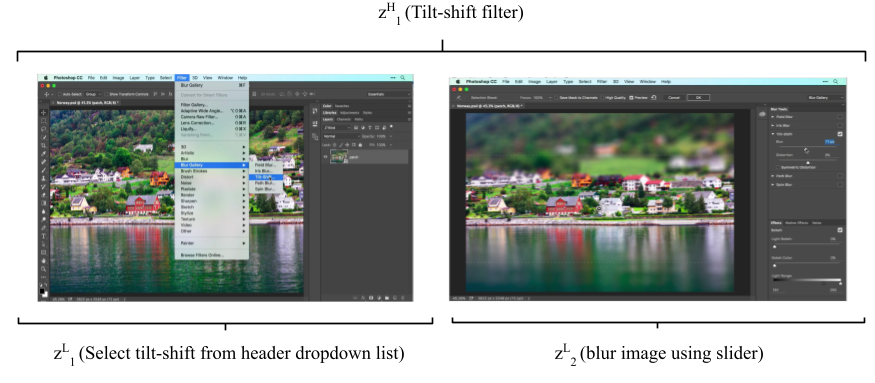}
    \caption{
 In this the high-level event \textit{Tilt-shift filter} is broken down into low-level events like \textit{Select tilt-shift from header dropdown list} and \textit{blur image using slider}.
    }
    \label{fig:ps5}
\end{figure*}
\begin{figure*}[t]
    \centering
    \includegraphics[width = 1.0\textwidth]{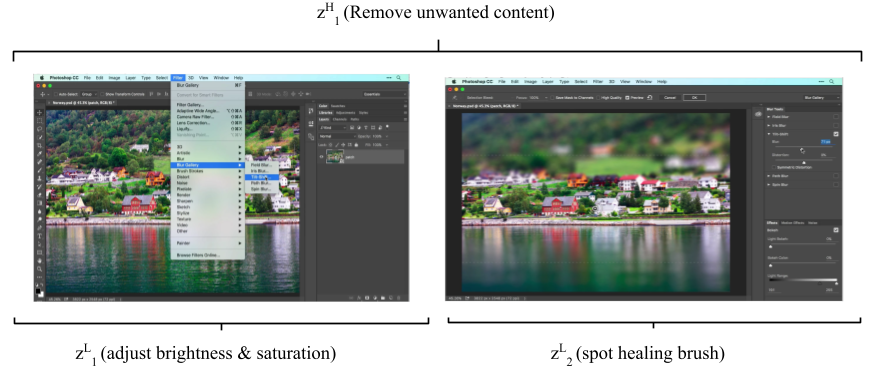}
    \caption{
 In this the high-level event \textit{Remove unwanted content} is broken down into low-level events like \textit{adjust brightness \& saturation} and \textit{spot healing brush}.
    }
    \label{fig:ps6}
\end{figure*}
\begin{figure*}[h]
    \centering
    \includegraphics[width = \textwidth,height=3cm]{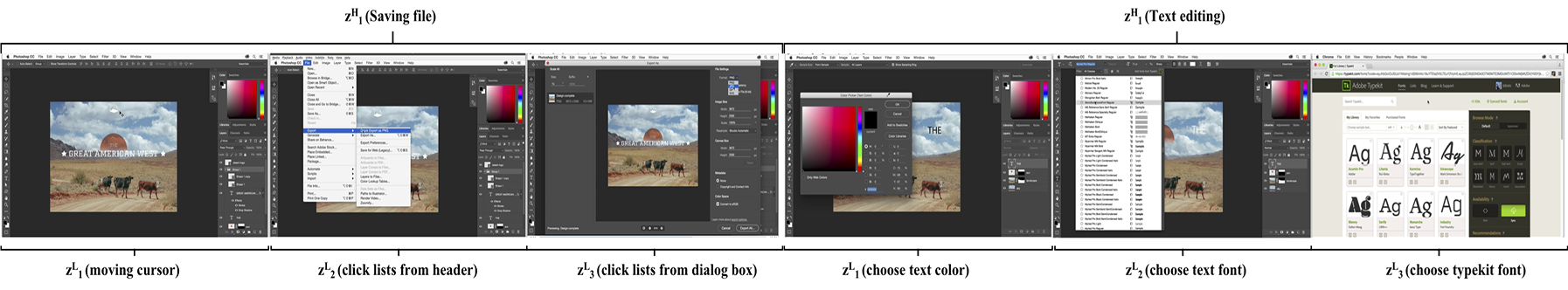}
    \caption{Example Hierarchy of events discovered by SHERLock on the TutorialVQA dataset}
    \label{fig:phototshop_hie}
\end{figure*}
\end{document}